\documentclass[11pt,a4paper]{article}
\pdfoutput=1

\usepackage[hyperref]{emnlp2020}

\usepackage{microtype}

\usepackage{times}
\usepackage{latexsym}

\usepackage{booktabs}
\usepackage{multicol,multirow}

\usepackage{graphicx}

\usepackage{todonotes}

\aclfinalcopy 

\title{The ELITR ECA Corpus}

\author{Philip Williams \and Barry Haddow\\
\\
  School of Informatics, University of Edinburgh, Scotland \\
  \texttt{\{pwillia4,bhaddow\}@ed.ac.uk}}
\date{}

\begin{document}
\maketitle
\begin{abstract}
We present the ELITR ECA corpus, a multilingual corpus derived from publications of the European Court of Auditors.
We use automatic translation together with Bleualign to identify parallel sentence pairs in all 506 translation directions.
The result is a corpus comprising 264k document pairs and 41.9M sentence pairs.
\end{abstract}

\section{Introduction}
Most machine translation systems are trained using large volumes of human-translated text.
The high cost of human translation means that training data is not typically created specifically for use in machine translation but is instead repurposed from sources where text has been translated for some other use.
The European Union, which publishes in up to 24 languages has long been a rich source of data for the machine translation community~\cite{Koehn2005,steinberger-etal-2006-jrc,10.1007/s10579-014-9277-0,hajlaoui-etal-2014-dcep}.

In line with other institutions of the EU, the European Court of Auditors\footnote{\url{https://www.eca.europa.eu/}} publishes the majority of its official documents in 23 of the 24 EU languages.\footnote{Irish is due to be included from 2022.}
To our knowledge, this data has not yet been compiled into a publicly available corpus.
In this work, we downloaded PDF reports from the ECA website and extracted plain text versions to create 23 monolingual corpora, with text segmented at the document and sentence level.
We then used a multilingual neural machine translation system to automatically translate text in all 506 translation directions and used Bleualign~\cite{sennrich-volk-2011-iterative} to identify parallel sentences.
The resulting dataset comprises 264k document pairs and totalling 41.9M sentence pairs (including duplicates) across all translation directions.
It is available from \url{http://data.statmt.org/elitr-eca}.

Sections~\ref{sec:monolingual} and ~\ref{sec:parallel} describe the methodology in more detail and report some further statistics of the data.

\section{Monolingual Corpora}
\label{sec:monolingual}
We created monolingual corpora in 23 languages by downloading PDF reports, extracting plain text, converting to UTF-8, splitting into sentences, de-hyphenating, and normalizing.

The PDFs were downloaded in May 2020.
We first used the ECA website's search facility\footnote{\url{https://www.eca.europa.eu/en/Pages/PublicationSearch.aspx}} to obtain a list of URLs for all available English language PDFs.
ECA documents use a consistent file naming convention where the ISO 639-1 language code is included in the filename (for example \texttt{agencies\_2019\_EN.pdf}).
We substituted the 22 other language codes and downloaded all documents that were available.

To obtain plain-text versions of the PDFs, we wrote an Automator script that used macOS's built-in support for PDF-to-text conversion.
The resulting files used a mixture of character encodings, so we used \texttt{chardet}\footnote{\url{https://chardet.github.io}} to automatically detect the encoding of each file, then converted from that encoding to UTF-8 using \texttt{iconv}.

We split the text into sentences using the \texttt{split-sentences.perl} script from the Moses toolkit~\cite{koehn-etal-2007-moses}.

A common problem with text extracted from PDF is that words that were broken at the end of a line retain their hyphenation and must be re-joined.
We used a simple statistical approach where we collected token counts for each language across all of the extracted text.
We then searched for pairs of tokens where the first ends in a hyphen, replacing the pair with their concatenation, either with or without the hyphen depending on which resulting form had been observed most frequently.
For example, \texttt{dur- ing} was replaced with \texttt{during} because \texttt{during} was seen 8,735 times and \texttt{dur-ing} was never seen.
\texttt{co- financed} was replaced with \texttt{co-financed} because that token was seen 1,079 times and \texttt{cofinanced} was seen only 63 times.

Finally, we normalized the Unicode representation files by converting them to NFC form.

Table~\ref{tab:monolingual} gives per-language statistics for the resulting corpus.

\begin{table}
    \centering
    \begin{tabular}{lrr}
    \toprule
    Language & Documents & Sentences \\
    \midrule
Bulgarian & 779 & 398,012 \\
Croatian & 444 & 281,783 \\
Czech & 871 & 418,043 \\
Danish & 368 & 193,895 \\
Dutch & 783 & 391,914 \\
English & 1,391 & 661,815 \\
Estonian & 402 & 183,265 \\
Finnish & 1,072 & 535,880 \\
French & 1,229 & 591,559 \\
German & 816 & 376,395 \\
Greek & 1,126 & 541,826 \\
Hungarian & 868 & 426,962 \\
Italian & 181 & 92,540 \\
Latvian& 198 & 94,864 \\
Lithuanian & 434 & 204,860 \\
Maltese & 867 & 398,056 \\
Polish & 867 & 412,476 \\
Portuguese & 1,130 & 527,610 \\
Romanian & 278 & 124,837 \\
Slovak & 869 & 434,025 \\
Slovene & 310 & 138,246 \\
Spanish & 1,143 & 555,153 \\
Swedish & 1,074 & 519,219 \\
    \bottomrule
    \end{tabular}
    \caption{Monolingual corpus statistics.}
    \label{tab:monolingual}
\end{table}

\section{Parallel Corpora}
\label{sec:parallel}
We derived parallel corpora from pairs of plain-text documents for all 506 directed combinations of the 23 languages.
For each plain-text version of a document $d_{x}$ in language $x$ with a corresponding plain-text document $d_{y}$ in language $y$, we translated $d_{x}$ into language $y$ to create document $d_{xy}$.
We then used Bleualign to identify parallel sentences in $d_{x}$ and $d_{y}$ based on the similarity of sentences in $d_{y}$ to the translations in $d_{xy}$.
We used a threshold similarity of 10.0 \textsc{Bleu}. We did this for each language pair $x$-$y$ to create a corpus of parallel sentences, creating separate corpora for pairs $x$-$y$ and $x$-$y$, from the same set of documents.
This provides the option of taking the intersection of sentence pairs obtained from translation in both directions, which should be expected to increase the precision of parallel sentence identification.

In order that the corpus can be filtered based on translation quality, we provide files in tab-separated format, one per document pair, containing the source sentence, target sentence, and \textsc{Bleu} score.

Table~\ref{tab:parallel} gives some statistics for the resulting corpus.
For space reasons, we show only the top 10 and bottom 10 language pairs, ordered by number of sentence pairs.

\begin{table}
    \centering
    \begin{tabular}{p{2cm}p{2cm}p{2cm}}
    \toprule
    Language Pair & Document Pairs & Sentence Pairs \\
    \midrule
    fr2en & 1,203 & 455,227 \\
en2fr & 1,203 & 454,616 \\
es2en & 1,124 & 437,505 \\
en2es & 1,124 & 437,174 \\
pt2en & 1,112 & 414,160 \\
en2pt & 1,112 & 409,990 \\
el2en & 1,104 & 404,410 \\
en2el & 1,104 & 400,482 \\
sv2en & 1,066 & 398,566 \\
en2sv & 1,066 & 391,925 \\
\ldots & \ldots & \ldots \\
it2lv & 135 & 14,923 \\
et2it & 134 & 14,915 \\
lt2it & 135 & 14,888 \\
lv2hr & 96 & 14,848 \\
it2hu & 134 & 14,497 \\
it2et & 134 & 14,348 \\
hu2it & 134 & 14,266 \\
bg2it & 122 & 13,973 \\
hr2it & 66 & 8,386 \\
it2hr & 66 & 8,300 \\
    \bottomrule
    \end{tabular}
    \caption{Parallel corpus statistics.}
    \label{tab:parallel}
\end{table}

\subsection{Translation Model}

For translation, we trained a single many-to-many multilingual model on a dataset sampled from the OPUS collection~\cite{TIEDEMANN12.463}.
Similar to prior work, including \citet{aharoni-etal-2019-massively} and \citet{ZhangEtAl2020}, our dataset was English-centric, meaning that all training pairs included English on either the source or target side.
We used all available parallel corpora for English paired with the 23 non-English EU languages.
In order to reduce the dominance of high-resource language pairs / domains, we applied exponentially smoothed weighting to domain and language pair choice when sampling sentence pairs.
In total, our training data contained 364M sentence pairs.

Our preprocessing pipeline consisted of corpus cleaning, segmentation, and target language tagging.
For corpus cleaning, we used the \texttt{clean-corpus-n.perl} script from the Moses toolkit~\cite{koehn-etal-2007-moses}.
This applies a maximum length threshold of 80 as well as removing empty sentences and sentence pairs with length ratios greater than 9:1.
For segmentation, we trained a single SentencePiece  model~\cite{kudo-richardson-2018-sentencepiece} with a vocabulary size of 64,000 BPE subwords and a vocabulary threshold of 50.
We added a token to each source sentence to specify the target language, as in~\citet{JohnsonEtAl2017}.

For training we used Marian~\citep{mariannmt} with hyperparameter settings matching the `base' configuration of~\citet{vaswani_attention_2017}.

\section{Conclusion}
We have presented the ELITR ECA corpus, a multilingual corpus derived from publications of the European Court of Auditors.
The corpus preserves document boundaries and \textsc{Bleu} similarity scores, and contains between 8,300 and 455,227 sentence pairs per language pair in all 506 translation directions, totalling 41.9M sentence pairs.

\section*{Acknowledgements}
This work has received funding from the European Union's Horizon 2020 Research and Innovation Programme under Grant Agreement No 825460 (ELITR).

\bibliography{references}
\bibliographystyle{acl_natbib}


\end{document}